\documentclass{llncs}

\usepackage{booktabs} 
\usepackage{subdepth}
\usepackage{bm}
\usepackage{dsfont}
\usepackage{xspace}
\usepackage{algorithm}
\usepackage{amsmath}
\usepackage{amsfonts}
\usepackage{graphicx}
\usepackage{cleveref}
\usepackage{pdfpages}
\usepackage[noend]{algpseudocode}
\usepackage{todonotes}
\usepackage{paralist}
\usepackage{pgf}
\usepackage{subfigure}
\usepackage{booktabs}

\begin{document}
\title{Distance-based Kernels for Surrogate Model-based Neuroevolution\thanks{This publication was accepted to the Developmental Neural Networks Workshop of the Parallel Problem Solving from Nature 2018 (PPSN XV) conference}}

\author{J\"org Stork, Martin Zaefferer, and Thomas Bartz-Beielstein}

\institute{
TH K\"oln, Institute for Data Science, Engineering, and Analytics, \\
Steinm\"ullerallee 1,~51643~Gummersbach, Germany, \\
\email{joerg.stork|martin.zaefferer|thomas.bartz-beielstein\\@th-koeln.de}}

\maketitle

\begin{abstract}
The topology optimization of artificial neural networks can be
particularly difficult if the fitness evaluations require expensive
experiments or simulations. For that reason, the optimization methods
may need to be supported by surrogate models. We propose different distances for
a suitable surrogate model, and compare them in a simple 
numerical test scenario.
\end{abstract}

\section{Introduction}
A crucial and difficult task in the optimization of Artificial Neural Networks (ANN)
is the optimization of their topology. Often, only parameters like
the edge weights, number of hidden layers, and number of elements
per layer are considered.
More fine-grained and less restrictive changes to the network structure 
are rarely examined. Two notable exceptions are 
NeuroEvolution of Augmenting Topologies (NEAT)~\cite{stanley2002evolving} 
and Cartesian Genetic Programming of Artificial Neural Networks (CGPANN)~\cite{turner2013cartesian}. 
Both NEAT and CGPANN require numerous fitness evaluations, which 
makes them inefficient for applications with expensive function evaluations such as simulations or real-world
experiments, e.g., when an ANN is the controller
of a robot that operates in a complex real-time environment.
In such cases, it would be desirable to employ surrogate models
that reduce the load of fitness evaluations.
While surrogate models are frequently employed in continuous 
optimization, more complex discrete search spaces are
less often investigated~\cite{Bart16n}. 
Recently, a first surrogate model for ANN optimization 
was applied to NEAT by Gaier et al.~\cite{Gaier2018a}.
They use a distance-based surrogate model, employing
a genotypic compatibility distance that is part of NEAT.
We will focus on investigating
the effect of different distances on the search performance
of a surrogate model, optimizing ANNs with CGPANN.
We compare distances that take the genotype (network structure)
and/or phenotype (behavior) of the ANNs in account.
A transparent and simple numerical test case is used to evaluate
the performance of the different distances.

\section{Surrogate Modeling for Neuroevolution}
We use a Surrogate Model-based Optimization (SMBO) algorithm, 
following the idea of Efficient Global Optimization (EGO)~\cite{Jones1998}.
In short, a Kriging model is constructed based on an initial candidate data set, 
which is used in each iteration to suggest new promising candidates by optimizing the expected improvement criterion.
Kriging approximates the data by modeling the similarities of samples
via kernels and distances.
We employ the Kriging implementation of the R package 
\texttt{CEGO}~\cite{CEGOv2.2.0,Zaefferer2014b}.
It uses distance-based kernels
to model data from structured, combinatorial search spaces.
We utilize different genotypic and phenotypic distance measures for ANNs,
see Section \ref{sec:kernels}.
For CGPANN, we use the C library \texttt{CGP} by A.~Turner\footnote{http://www.cgplibrary.co.uk - accessed: 2018-01-12},
which also allows the application of neuroevolution~\cite{turner2015introducing}. 
\texttt{CGP} was modified with function interfaces to R, distance measures, and fitness functions. 
The genotype of a CGPANN individual consists of a fixed number of nodes. 
Each node has a number of connection genes based on the pre-defined arity with adjacent weight genes and a single categorical function gene. 
Nodes are connected to preceding nodes. Duplicate connections to nodes are possible. 
Moreover, each node has a Boolean activity gene, which signals if it is used in the active phenotype. 
CGPANN utilizes a (1+4)-Evolution Strategy.

\section{Proposed Kernels and Distances}\label{sec:kernels}
In the following, we always use an exponential kernel $k(x,x')=\exp(-\theta d(x,x'))$.
Here, $x$ and $x'$ represent the optimized ANNs, and consist of 
the weights $x_{w}$, input labels $x_i$, activity labels $x_a$, and transfer function labels $x_f$, i.e.,
$x=\{x_w,x_i,x_a,x_f\}$.
All distances $d(x,x')$ are scaled to $[0,1]$.  \\
\emph{Genotypic Distance (GD):} 
The simple genotypic distance combines the distances of weights, inputs, activity of nodes, and transfer functions. 
The inputs are sorted for each $x$ before computation of the distance $d(x,x')= ||x_w-x_w'||^2+ H(x_i,x_i') +H(x_a,x_a') +H(x_f,x_f')$, where
$H(a,b)$ denotes the Hamming distance, i.e., the transfer function distance is 0 if the function is identical, else 1. \\
\emph{Genotypic ID Distance (GIDD):} 
A distance based on the active topology,  utilizing IDs for each node. 
Each active node in the ANN is given an ID based on the connections to prior nodes or inputs and the number of non-duplicate 
connections of this node. 
The distance is calculated by a pairwise comparison of all node IDs. 
If a node ID is present in both ANNs, the  subgraph prior to this node is analyzed recursively. 
If all IDs in the subgraph also match, we assume that the corresponding nodes have an equal position in the ANN structure. 
For all nodes that are matched in this way, the Euclidean distance of the weights ($x_w$) and Hamming distance of
the transfer functions ($x_f$) is computed.
A node pair can only be used once for this comparison, as node IDs may be present several times in each individual. 
If all node IDs of both individuals $x$ and $x'$ are equal, the GIDD is simply the distance between
all weights and transfer functions. \\
\emph{Phenotypic Distance (PD):} 
A data set is used to compute the outputs of each ANN. 
For each sample, the \emph{softmax} of the outputs is computed, which yields the class probabilities.
The PD is the Euclidean distance of all softmax values.
Different data sets can be created, e.g., based on a subset of the original task data set.
Else, a distinct data set can be created by design of experiment methods. 

\noindent
We also considered employing the graph edit distance, but decided against it due to its complexity. Computing the graph edit distance is NP-hard~\cite{Zeng2009}.

\section{Experiments}
\emph{Benchmark Task and Setup:}
The benchmark task is the neuroevolution of an ANN with a small budget of 250 total evaluations, which provides a realistic scenario for problems with expensive fitness evaluations. 
The well-known IRIS data set ($n=150$ samples, 4 variables, and 3 classes), is used as an elementary benchmark problem. 
The fitness function is the adjusted classification accuracy: 
$\text{acc}=\sum_{i=1}^n a_i$, where $a_i=1$ if the predicted class is true, otherwise, $a_i$ is the predicted probability for
the true class.
In this preliminary study, the data set was not partitioned into train and test data. 
As baselines, we used random search and the original CGPANN with different mutation rates. 
Moreover, the above described distance measures are compared. For the PD, three sample methods are used: the complete data set (baseline) and two Response Surface Method designs (RSM) based on the IRIS parameter boundaries with a small (15) and a large (60) sample size. 
In SMBO, the (1+4)-ES is used in each consecutive iteration alternating between exploiting (L) and exploring (G). 
Parameters are listed in Table~\ref{tab:setup}. Algorithm parameters were not tuned.\\
\begin{table}[b]
\centering
\caption{Parameter setup, where evaluations denote the initial candidates plus the budget for consecutive evaluations. 40 Nodes were used to keep the search space small, but sufficient for the IRIS problem given this small budget.}
\label{tab:setup}
\begin{tabular}{l|l|l|l}
  \bf arity & \bf nodes & \bf weight range & \bf function set \\
  5 & 40 &  {[}-1,1{]} & \multicolumn{1}{l}{ tanh, softsign, step, sigmoid, gauss}\\
\midrule
\bf method  & \bf mutation rate & \bf evaluations & \bf surrogate evaluations \\
Random &   & 250 &  \\
CGPANN &  5\%/15\% &  $1+4\cdot63$&   \\
SMBO &  L:5\% G:15\% & 50+200 & L:10+400 G:1000+400\\
\end{tabular}
\end{table}
\emph{Results and Discussion:}
\begin{figure}[ht]
\centering
\includegraphics[width=.95\textwidth]{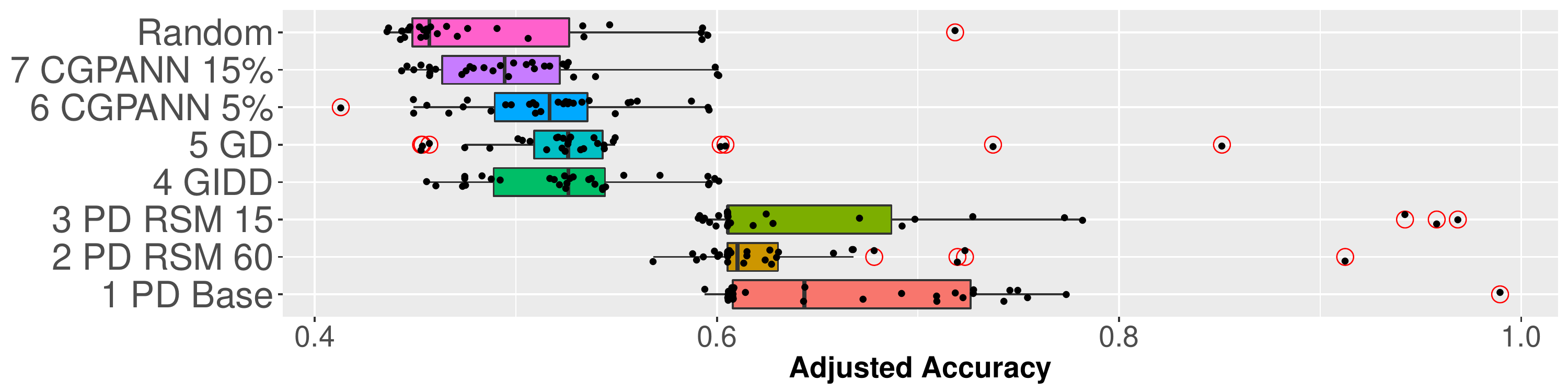}
\caption{Results after 250 fitness evaluations with 30 replications, comparing random search, CGPANN and different SMBO-CGP variants. 
Red circles depict outliers.}
\label{fig:results}
\end{figure}
Figure \ref{fig:results} visualizes the results. 
With the given small number of evaluations, CGPANN obtained a better median value than random search. 
The genotypic distances performed similarly to CGPANN, while the PD distances performed better. 
An analysis of the convergence (not shown here) indicates a tendency to end up at a local optimum around an accuracy of 66\%, 
which can be explained by the underlying benchmark problem with three classes. 
Importantly, even the small RSM set performs only slightly worse than the base set. 
The poor performance of GD and GIDD can be explained by the fact that even smallest changes in the genes can have a large impact on the ANN output. 
In contrast, the PD distances directly exploit the ANN behavior.

\section{Conclusion and Outlook}
In this work, we have shown that SMBO is a promising extension for CGPANN, that is able to improve the neuroevolution of ANNs in case of small evaluation budgets. 
SMBO with phenotypic distance kernels, which are based on generated outputs of each ANN, shows significantly better results than with genotypic distances or basic CGPANN. In future work we will investigate a larger set of tasks, generalization ability, longer runs, parameter tuning, and a more thorough investigation of the promising phenotypic distance.

\noindent
{\footnotesize {\textbf{Acknowledgements:}} This work is part of a project that has received funding from the European Union's Horizon 2020 research and innovation program under grant agreement no. 692286.}

\bibliographystyle{splncs}
\bibliography{Stor18a}  

\end{document}